\documentclass[10pt,twocolumn,letterpaper]{article}

\usepackage[pagenumbers]{cvpr} 

\usepackage{graphicx}
\usepackage{amsmath}
\usepackage{amssymb}
\usepackage{booktabs}
\usepackage{multirow}
\usepackage{soul}
\usepackage{subcaption}
\usepackage[table,usenames,dvipsnames]{xcolor}
\usepackage{pifont}
\usepackage{pgfplots}

\definecolor{DarkPurple}{RGB}{17, 36, 107}
\definecolor{DarkBlue}{RGB}{15, 25, 76}
\definecolor{DarkBlue2}{RGB}{27, 98, 117}
\definecolor{DarkGreen}{RGB}{51, 76, 73}
\definecolor{DarkYellow}{RGB}{167, 172, 131}
\definecolor{Purple}{RGB}{132, 137, 185}
\definecolor{LightBlue}{RGB}{95, 173, 209}
\definecolor{LightGreen}{RGB}{168, 228, 170}
\definecolor{LightYellow}{RGB}{245, 236, 154}
\definecolor{Gray}{gray}{0.88}
\definecolor{DarkGray}{gray}{0.8}
\definecolor{LightGray}{gray}{0.93}
\pgfplotsset{compat=1.8}

\usepgfplotslibrary{statistics}

\usepackage[pagebackref,breaklinks,colorlinks]{hyperref}

\usepackage[capitalize]{cleveref}
\crefname{section}{Sec.}{Secs.}
\Crefname{section}{Section}{Sections}
\Crefname{table}{Table}{Tables}
\crefname{table}{Tab.}{Tabs.}

\mathchardef\mhyphen="2D
\def\cvprPaperID{16} 
\def\confName{CVPR}
\def\confYear{2023}

\begin{document}

\title{Understanding the Challenges and Opportunities of Pose-based Anomaly Detection}
\author{Ghazal Alinezhad~Noghre\textsuperscript{1}\thanks{Corresponding author.} \\
\and
Armin Danesh~Pazho \textsuperscript{1}\\
\and
Vinit Katariya\textsuperscript{1}\\
\and 
Hamed Tabkhi\textsuperscript{1}\\
\and 
\textsuperscript{1} University of North Carolina at Charlotte\\
\texttt{\{galinezh, adaneshp, vkatariy, htabkhiv\}@uncc.edu}
}
\maketitle

\begin{abstract}
Pose-based anomaly detection is a video-analysis technique for detecting anomalous events or behaviors by examining human pose extracted from the video frames. Utilizing pose data alleviates privacy and ethical issues. Also, computation-wise, the complexity of pose-based models is lower than pixel-based approaches. However, it introduces more challenges, such as noisy skeleton data, losing important pixel information, and not having enriched enough features. These problems are exacerbated by a lack of anomaly detection datasets that are good enough representatives of real-world scenarios. In this work, we analyze and quantify the characteristics of two well-known video anomaly datasets to better understand the difficulties of pose-based anomaly detection. We take a step forward, exploring the discriminating power of pose and trajectory for video anomaly detection and their effectiveness based on context. We believe these experiments are beneficial for a better comprehension of pose-based anomaly detection and the datasets currently available. This will aid researchers in tackling the task of anomaly detection with a more lucid perspective, accelerating the development of robust models with better performance.

\end{abstract}


\section{Introduction}
Video Anomaly Detection (VAD) has recently become one of the trending topics among researchers. It focuses on identifying abnormal behaviors and distinguishing them from normal behaviors. VAD is shown to be beneficial in different applications and fields, including smart video surveillance \cite{pazho2023ancilia, sultani2018real, patrikar2022anomaly}, crowd control and analysis \cite{rezaee2021survey, khan2022crowd}, industrial inspection \cite{liu2021semi, lu2022deep}, and many more.

VAD is inherently context-specific, and anomalous behaviors in one context may be considered normal in another environment. In addition, the anomaly detection problem is an open-set problem, meaning that behaviors can happen in various ways that might not be accounted for in the datasets. It requires a robust model to detect new and unseen behaviors and generalize over the new data. Translation from one domain (i.e., dataset) to another often significantly impacts the models' performance.

One of the branches of video anomaly detection that identifies whether a person's behavior is anomalous is called Person Anomaly Detection (PAD). There are two primary techniques to carry out PAD. One is based on the pixels data, and one is based on a person's skeletal information (pose). In general, pixel-based algorithms \cite{wu2022self, tian2021weakly} deliver improved results compared to pose-based approaches \cite{stg_nf, markovitz2020graph, morais2019learning}. However, from an ethical and privacy standpoint, research has demonstrated that pose-based algorithms are superior to their alternative \cite{ardabili2022understanding}. They tend to remove biases such as gender, race, and age. Computationally, pose-based approaches are less complex, as the extracted pose representation is concise and has fewer data points than pixel-based analysis. They have attracted much attention and are being adopted in various applications \cite{pazho2023ancilia}. However, several complex challenges hinder their effectiveness.

Adopting pose-based anomaly detection in the real world comes with some limitations. They only have access to noisy machine-generated keypoints. Occlusion, lighting, camera angle, and clothing variability affect the pose estimation, adding to the complexity. Understanding the intrinsic features of the available datasets is very useful for building more robust algorithms. It aids with understanding why some models work perfectly on one dataset, producing State-of-the-Art (SotA) results. However, they show suboptimal performance when trained and validated on other datasets.

In this work, we adopt pose information of people to analyze the major available datasets and explore their characteristics. In addition, we use the trajectory of the people for further investigation. Trajectory and its prediction have been highly explored in various fields \cite{noghre2022pishgu, salzmann2020trajectron++, lin2021vehicle, katariya2022deeptrack}, and it is shown to be beneficial for a wide range of tasks including anomaly detection. We conduct various experiments based on pose and trajectory to understand why one dataset is more challenging than another and why algorithms that work perfectly on one might not work accurately on another. All the contributions of this article aim to quantify the difficulty and challenges of two primary VAD datasets and the discriminative capabilities of pose and trajectory. In summary, the contributions are:

\begin{itemize}

    \item Analyzing and comparing SotA pose-based approaches on existing anomaly detection datasets and evaluating the impacts of two new orthogonal feature representations (single-person trajectory and social trajectories) to enhance their discriminative power for anomaly detection. (Section \ref{sec:compare}, \ref{sec:pose_normal}, \ref{sec:traj_normal})
    
    \item Introducing a new metric, Signed Difference of Means ($S\mhyphen DoM$), to extract the inherent discriminative power of datasets based on trajectory and pose. (Section \ref{sec:sdom}) 
    
    \item Evaluating the difficulty of anomaly detection datasets using our newly introduced $S\mhyphen DoM$ metric to capture discriminative power of pose and trajectory per dataset. (Section \ref{sec:mean_analysis})
\end{itemize}

\section{Related Work}

Anomaly detection is a method of detecting behaviors or data points significantly different from the "normal" or "typical" behavior. The normal behavior is the expected data pattern observed across most data points. Traditionally, trajectory-based approaches \cite{smolyak2020coupled, bouritsas2019automated} were explored for detecting anomalies by observing human movements and learning common patterns. However, contemporary video-based methods \cite{zaheer2022generative,hao2022spatiotemporal} discussed below were found to be more accurate, utilizing deep learning techniques.

Numerous methods have been proposed for detecting anomalous behaviors using video-based anomaly datasets. These datasets are videos of normal and abnormal behaviors in the training and validation set. One of the most recent datasets, Charlotte Anomaly Dataset (CHAD) \cite{pazho2022chad}, provides high-resolution multi-camera anomaly detection with bounding box, identity, and pose annotations. ShanghaiTech Campus dataset \cite{lou2021ShanghaiTechAnomaly} is a well-known dataset that provides a collection of 317K normal and anomalous frames across several scenes. IITB-Corridor \cite{Royston2020IITB_Corridor} has one of the most extensive collections of frames but focuses on a single location which may not be generalized to other settings.

In contrast with previously mentioned datasets which collect videos from continuous streaming cameras, UCF Crime \cite{Sultani2018UCF_Crime} is based on the videos collected from YouTube, covering many scenes and thus changing the context with each scene. This may not be ideal for anomaly detection as it is context-specific. Another new dataset is the NOLA dataset\cite{Doshi2022NOLA}, which consists of a single scene covered by a moving camera. One of the characteristics of NOLA is that they use a moving camera, making them unique compared to their peers.

Unsupervised learning approaches for vision-based anomaly detection are well-established and widely used. They are trained only using normal videos and are able to detect anomalies as they are drastically different from normal behavior. Ongoing research in this field has resulted in sophisticated methods that are highly accurate and effective in different environments. Normal Graph \cite{luo2021normal} utilizes a spatiotemporal graph convolutional networks-based prediction approach and the structure of graphs at skeletal keypoint joints to detect outliers on the ShanghaiTech dataset. Spatiotemporal Graph Convolutional Autoencoder with Embedded Long Short-Term Memory Network (STGCAE-LSTM) \cite{li2022human} as the name suggests, uses a combination of graph autoencoders and LSTM network with dual-encoders to detect anomalies. The dual-encoder combines the reconstruction and prediction of the input to calculate a final anomaly score. Message-Passing Encoder-Decoder Recurrent Network (MPED-RNN) \cite{morais2019learning} separate skeletal movements into a global movement and local postures, and capture the anomaly using these separated features. Liu et al. introduce Self-Attention Augmented Spatiotemporal Graph Convolutional network (SAA-STGCN) \cite{liu2022self}, which first extracts the poses and uses the encoder of SAA-STGCAE followed by the Dirichlet process mixture model for generating a normality score to find the outliers. Spatio-Temporal Graph Normalizing Flows (STG-NF) \cite{hirschorn2022normalizing} similar to previously discussed methods works on time sequence data of human skeleton poses to build a spatiotemporal graph. Next, the algorithm utilizes generative models that normalize complex distributions and learn the bijective mapping between skeletal pose and Gaussian distribution. Graph Embedded Pose Clustering for Anomaly Detection (GEPC) \cite{markovitz2020graph} is used to detect anomalous behaviors using pose-based graphs mapped to latent space and bundled together to compare with normal actions.

\section{Preliminaries}

\subsection{Anomaly Detection Datasets}
In this work, we use the \emph{ShanghaiTech campus dataset} \cite{lou2021ShanghaiTechAnomaly} and recently introduced \emph{CHAD} \cite{pazho2022chad}. ShanghaiTech is widely used for pose-anomaly detection since it provides videos with good enough quality for pose extraction. The CHAD dataset provides official pose data and defines a new benchmark for pose anomaly detection. Hence, we choose to analyze these two datasets in this work. 

\textbf{ShanghaiTech Campus} dataset is a collection of approximately 270K training and 42K validation frames. It contains 17K anomalous frames in the validation set. These numbers add up to approximately 317K frames spread across 13 scenes recorded at a university campus. Various camera angles and different lighting conditions add to this dataset's versatility. ShanghaiTech does not contain pose annotations. Multiple works \cite{stg_nf, markovitz2020graph} extract pose information with AlphaPose \cite{alphapose}, and we obtained them from the STG-NF repository.

\textbf{CHAD} provides a multi-camera view of a parking lot scene with 1.15 million high-resolution frames at 30 frames per second, dividing into 1.02 million and 126K of training and validation frames. It includes pose annotations (keypoints), bounding box information, identity, and frame-level anomaly labels. It comprises 59K anomalous frames with 22 types of anomalous behaviors divided between individual and group anomalies.

\subsection{Pose-based Anomaly Detection Approaches}

We focus on unsupervised approaches because of the nature of VAD. Avidan \emph{et al.'s} GEPC and STG-NF stand out with SotA results on multiple anomaly benchmarks and the availability of the implementation of their work.

\textbf{GEPC} uses pose-based graphs with each keypoints as a graph node, which the spatiotemporal graph auto-encoder uses for automatic feature learning. The auto-encoder embeddings are exploited in the clustering stage to similar group actions in the latent space and assign a probability to each representation based on the cluster. In the final stage, GEPC utilizes a trained Dirichlet process mixture model to evaluate the normal behavior cluster distribution, generating a log probability as a sample score in the testing stage.

Recently introduced \textbf{STG-NF} is a lightweight PAD model that uses normalizing flows to handle unique characteristics of the graph-based pose data. It learns the mapping between input pose-data distribution to the latent Gaussian space distribution. The architecture of STG-NF uses a convolutional graph network. It has K flow steps, each divided into an activation normalizing layer, a permutation layer for reversible permutations using a linear transformation, and affine coupling layers using GCN to extract the relevant information from pose data. The pose sequence probability is the final normality score. In this study, we focus on the unsupervised setting of STG-NF.

\subsection{Evaluation Metrics}
Most PAD models use Area Under the Receiver Operating Characteristic (AUC-ROC) based on evaluation methodology proposed by Lou \emph{et al.}\cite{luo2017revisit}. However, Pazho \emph{et al.} argue that AUC-ROC information may not be enough to evaluate PAD models' performance. Using Area Under the Precision-Recall Curve (AUC-PR) and Equal Error Rate (EER) can provide additional information to understand the more profound insights into the model's performance.

\textbf{AUC-ROC} is used for binary classification and obtained by measuring the area under the curve for plotting the true-positive rate against the false-positive rate over different thresholds. A higher AUC-ROC usually indicates whether the model can classify respective classes well \cite{luo2017revisit}.

\textbf{AUC-PR} is the area under the curve when plotting precision against the recall. It provides further understanding of the model's capabilities and accounts for false-negative classifications while focusing on the positive class\cite{saito2015precision}, which is suitable for imbalanced datasets\cite{ma2013imbalanced}.

\textbf{EER} is calculated by plotting the model's false-negative and false-positive rates for various thresholds and finding the intersection point of the two curves. This indicates the exact threshold value that balances false-negative and false-positive classifications\cite{li2013anomaly}. It provides additional information to the insights obtained from AUC-ROC and AUC-PR to better understand the model performance. 

\subsection{Comparison between Existing Approaches}

\label{sec:compare}

\begin{figure*}[t!]
    \centering
    \resizebox{0.9\linewidth}{!}{
    \includegraphics[clip,trim={18 22 20 18},width=0.95\textwidth]{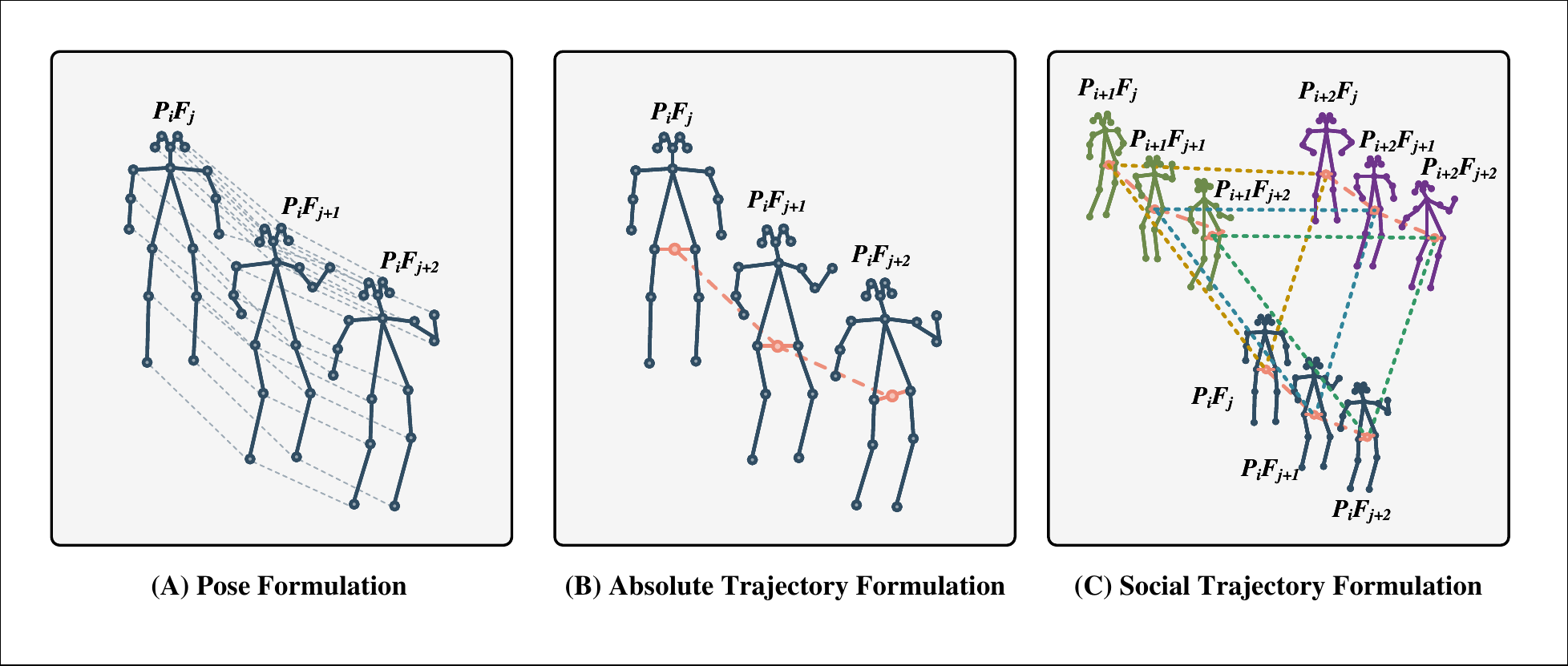}
    }
    \vspace{-10pt}
    \caption{Different input formulation used in our experiments. $P_iF_j$ represents the $i^{th}$ person in frame $j$. We are just showing the sequences for three frames but the length of the sequences is set to $T$. (A): Pose sequences of a length $T$ of an individual person are collected to build a spatiotemporal input graph. (B): Center of the right hip and left hip represents a person's location. The connection of $T$ centers builds the input sequence. (C) The location of all people available in the scene is used to construct a fully connected graph in each frame. This graph is also collected for T frames to build a spatiotemporal social trajectory graph.}
    \label{fig:formulation}
\end{figure*}

STG-NF \cite{stg_nf} in the unsupervised setting has been trained and evaluated on only the ShanghaiTech dataset \cite{lou2021ShanghaiTechAnomaly}. The authors provide the trained model, which we use to report AUC-PR and EER in addition to AUC-ROC. Intending to evaluate the adaptability of STG-NF to different datasets, we train and evaluate it on CHAD \cite{pazho2022chad} as well, using the same parameters and the setup suggested by the authors. The learning rate and the weight decay are set to $5e^{-4}$ and $5e^{-5}$, respectively, and the model is trained with the batch size of 256 for eight epochs using the Adamax optimizer. The results can be seen in Table \ref{tab:compare}. Although STG-NF has SotA results on ShanghaiTech by a far margin ($11.8\%$, $21.8\%$ and 0.09 improvement in AUC-ROC, AUC-PR, and EER respectively compared to GEPC), it does not show improved results on CHAD. STG-NF exhibits a $4.3\%$ drop in AUC-ROC and $0.05$ increase in EER compared to GEPC. This drop in performance is caused by the different characteristics of the two datasets and how different models can adapt to each of them. To better understand these datasets' properties and challenges, we conduct detailed tests in the following sections and quantify the discriminative power of different features in these two datasets. 

\begin{table}[b]
\centering
\vspace{-10pt}
\caption{Comparison between STG-NF \cite{stg_nf} and GEPC \cite{markovitz2020graph} on ShanghaiTech \cite{lou2021ShanghaiTechAnomaly} and CHAD \cite{pazho2022chad} anomaly datasets. Results of GEPC are provided in \cite{pazho2022chad}.}
\vspace{-8pt}
\label{tab:compare}
\resizebox{\columnwidth}{!}{%
\begin{tabular}{c||cc|cc}
\rowcolor{DarkGray}
 & \multicolumn{2}{c|}{ShanghaiTech \cite{lou2021ShanghaiTechAnomaly}} & \multicolumn{2}{c}{CHAD \cite{pazho2022chad}} \\ \hline \hline
 \rowcolor{LightGray}
  & \multicolumn{1}{c|}{\begin{tabular}[c]{@{}c@{}}STG-NF\\ \cite{stg_nf}\end{tabular}} & \begin{tabular}[c]{@{}c@{}}GEPC\\ \cite{markovitz2020graph}\end{tabular} & \multicolumn{1}{c|}{\begin{tabular}[c]{@{}c@{}}STG-NF\\ \cite{stg_nf}\end{tabular}} & \begin{tabular}[c]{@{}c@{}}GEPC\\ 
\cite{markovitz2020graph}\end{tabular} \\ \cline{1-5} 
AUC-ROC(\%) $\uparrow$& \multicolumn{1}{c|}{85.9} & 74.1 & \multicolumn{1}{c|}{60.6} & 64.9 \\
AUC-PR(\%) $\uparrow$& \multicolumn{1}{c|}{87.5} & 65.7 & \multicolumn{1}{c|}{66.8} & 58.7 \\
EER $\downarrow$ & \multicolumn{1}{c|}{0.22} & 0.31 & \multicolumn{1}{c|}{0.43} & 0.38
\end{tabular}
}
\end{table}

\section{Embedded Pose Distribution}
\label{sec:pose_normal}
\begin{figure*}[htbp]
\centering
\begin{subfigure}{0.48\textwidth}
  \centering
  \includegraphics[width=\linewidth]{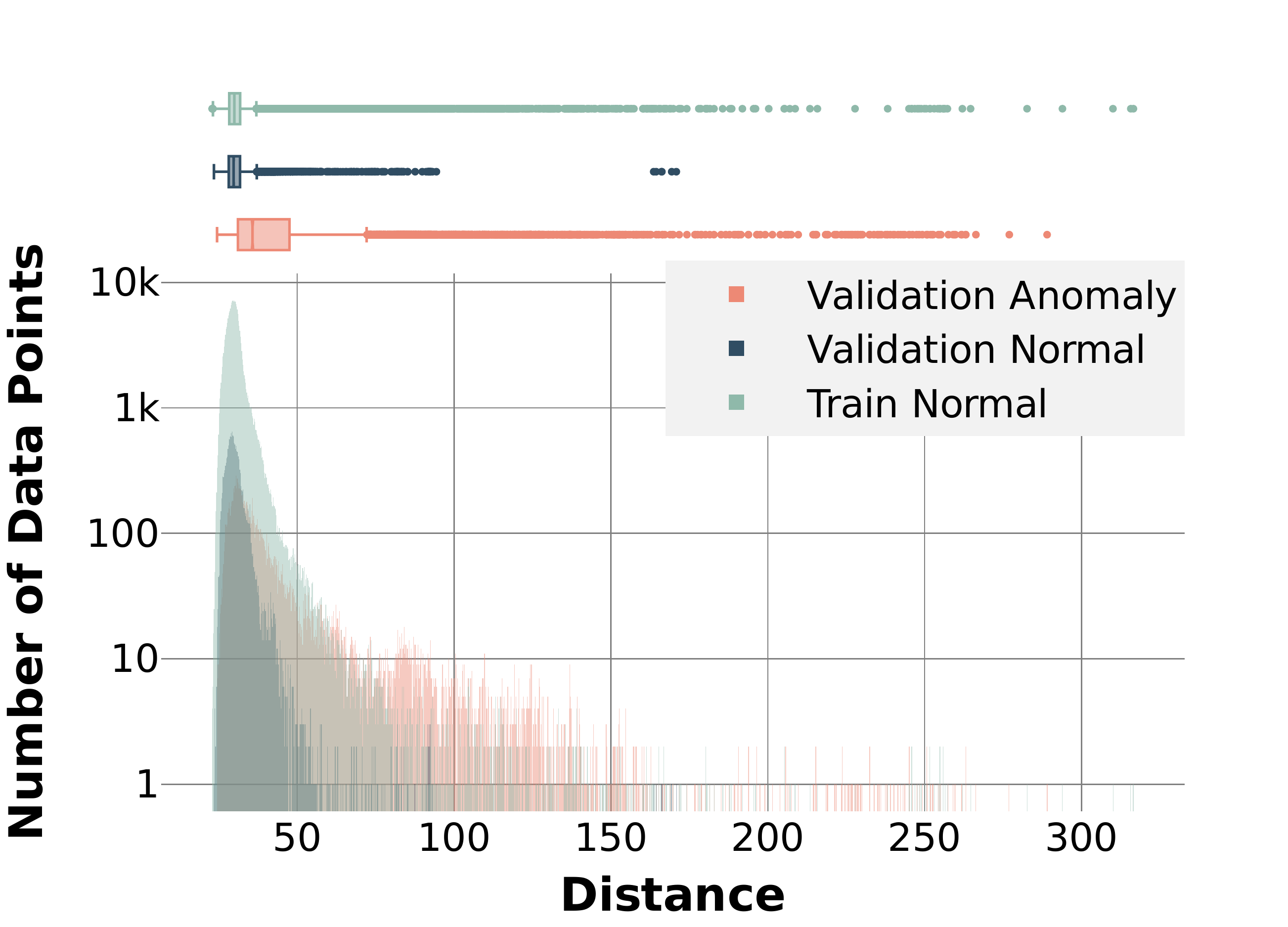}
  \caption{ShanghaiTech \cite{lou2021ShanghaiTechAnomaly}}
  \label{fig:pose_normal_shang}
\end{subfigure}%
\begin{subfigure}{0.48\textwidth}
  \centering
  \includegraphics[width=\linewidth]{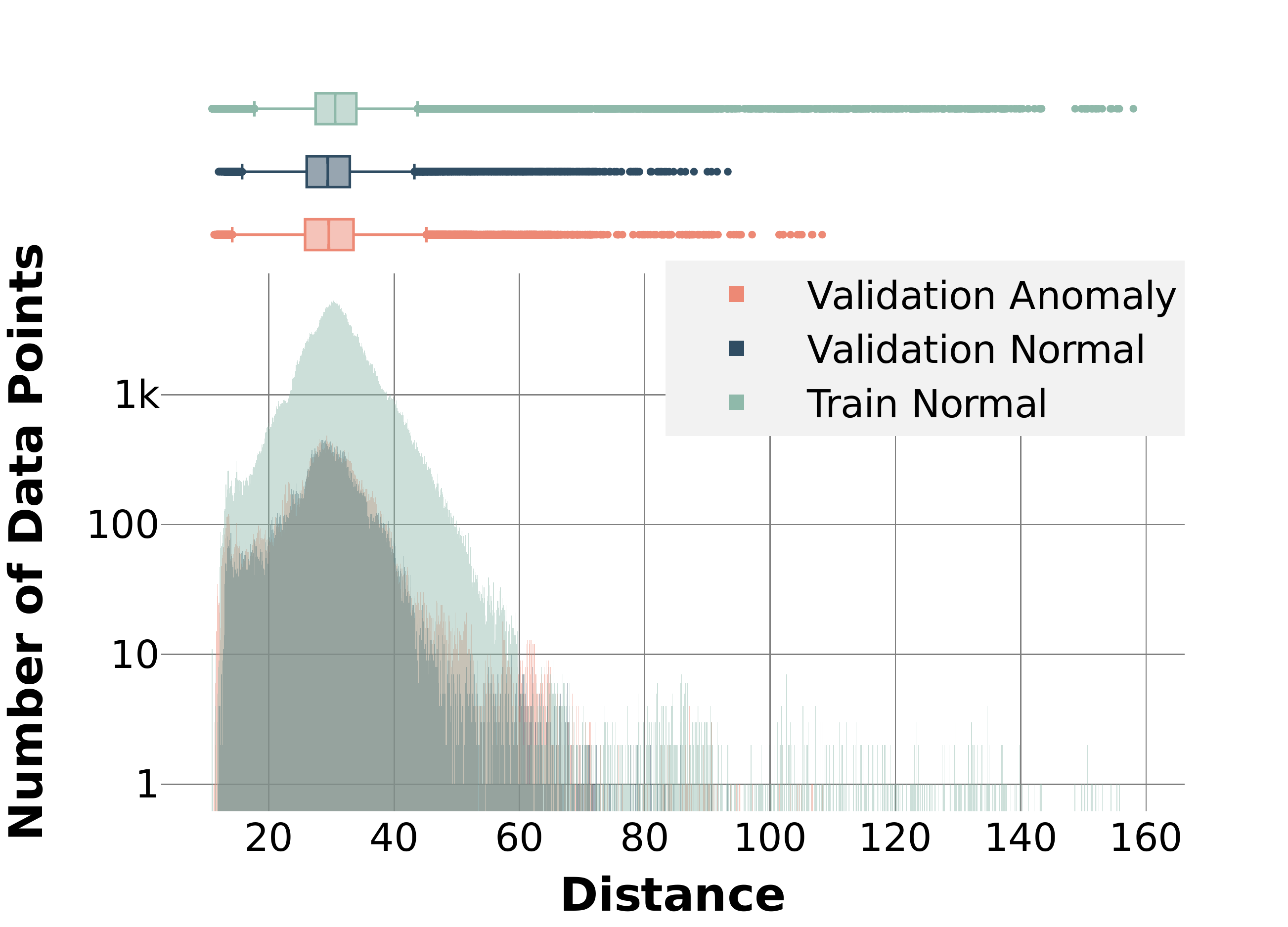}
  \caption{CHAD \cite{pazho2022chad}}
  \label{fig:pose_normal_chad}
\end{subfigure}%
\vspace{-8pt}
\caption{Histogram of the distances of all embedded pose sequences from the mean of the defined normal distribution ($\mu_{Normal}$) in the latent space. Detailed statistics can be found in Table \ref{tab:box}.}
\label{fig:pose_normal}
\end{figure*}

\begin{figure*}[htbp]
\centering
\begin{subfigure}{0.48\textwidth}
  \centering
  \includegraphics[width=\linewidth]{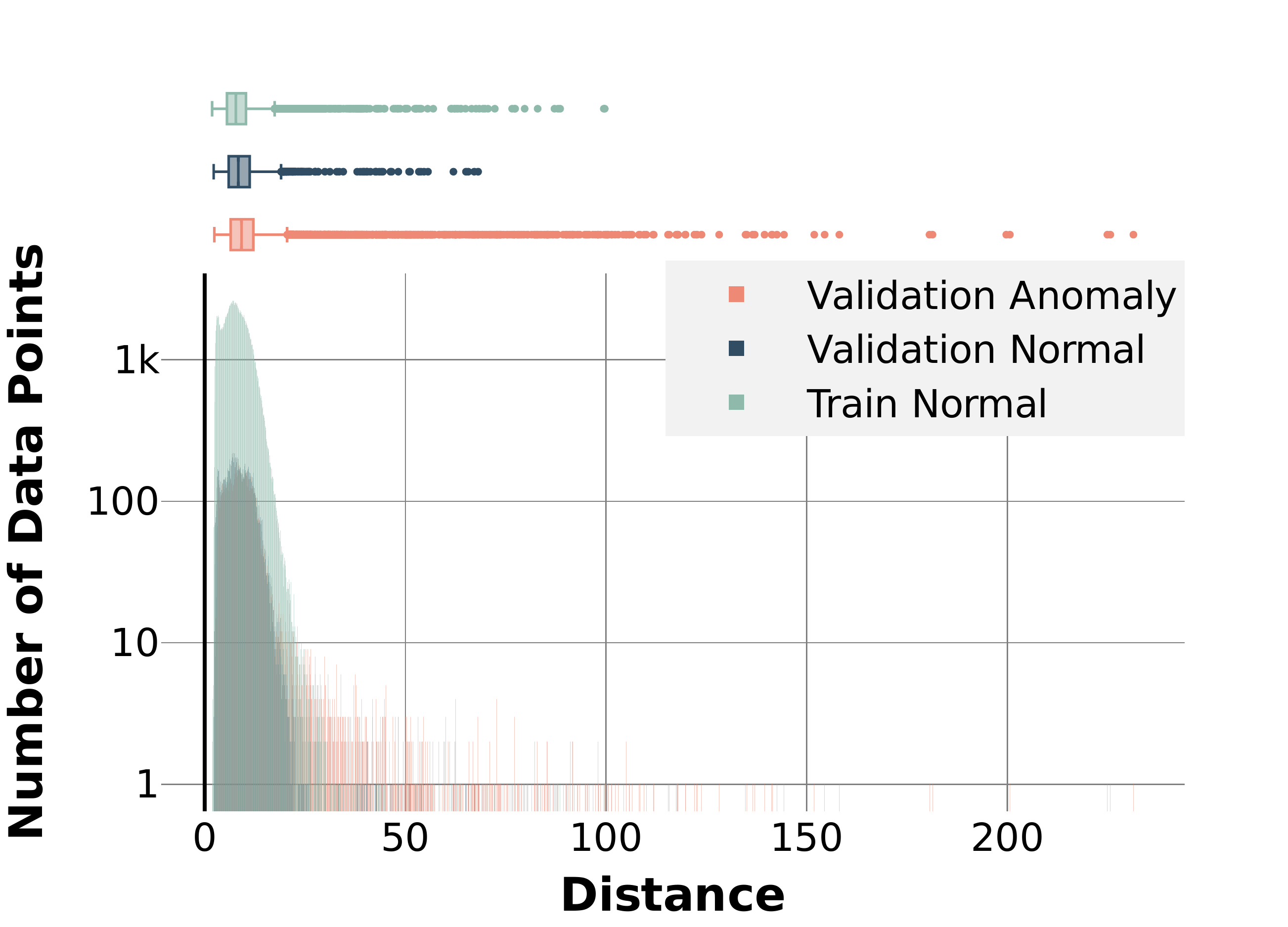}
  \caption{ShanghaiTech \cite{lou2021ShanghaiTechAnomaly}}
  \label{fig:traj_normal_shang}
\end{subfigure}%
\begin{subfigure}{0.48\textwidth}
  \centering
  \includegraphics[width=\linewidth]{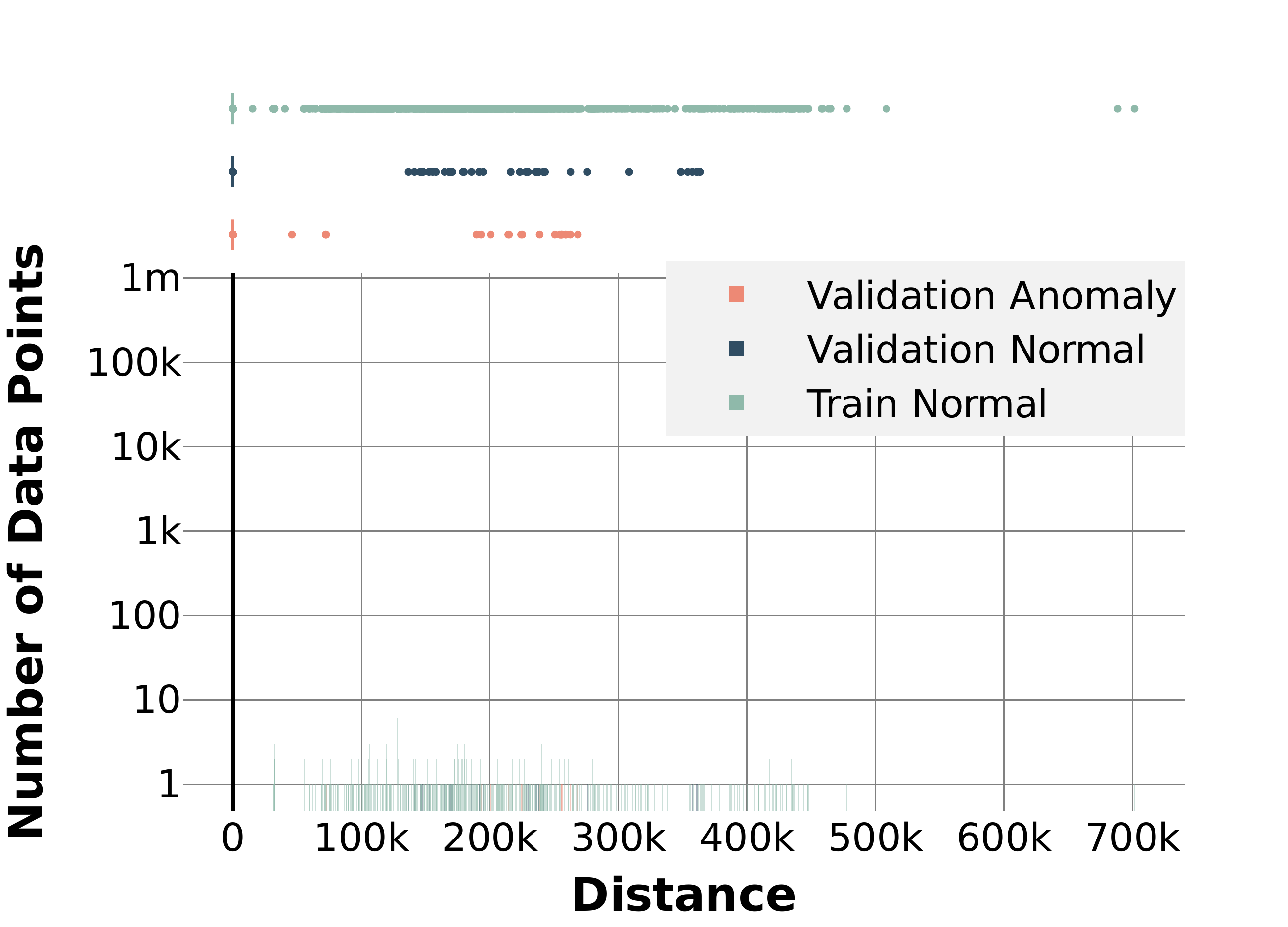}
  \caption{CHAD \cite{pazho2022chad}}
  \label{fig:traj_normal_chad}
\end{subfigure}%
\vspace{-8pt}
\caption{Histogram of the distances of all embedded absolute trajectories from the mean of the defined normal distribution ($\mu_{Normal}$) in the latent space. Detailed statistics can be found in Table \ref{tab:box}.}
\label{fig:traj_normal}
\end{figure*}

In this test, we use trained STG-NF \cite{stg_nf} anomaly detection models from Section \ref{sec:compare}, and map the input pose sequences to the latent space. The performance of these models is reported in Table \ref{tab:compare}. The formulation of the spatiotemporal input graph can be seen in Figure \ref{fig:formulation}. We replicated the experimental setup of \cite{stg_nf}. The input sequence length ($T$) is set to 24, and the stride is set to 6 frames. As a result, the network maps the input pose sequences to the latent space with a distribution of $Normal\backsim(\mu_{Normal},\Sigma)$. $\mu_{Normal}$ is the mean vector and $\Sigma$ is the covariance matrix of the normal distribution. We find the Euclidean distance between the embedded representation of the datapoints in both the training set and the validation set from $\mu_{Normal}$ using Equation \ref{eq:dl}.
\begin{equation}
\label{eq:dl}
    DL_i = ||L_{STG\mhyphen NF}(V^{T\times k\times 2}_i)-\mu_{Normal})||
\end{equation}

Where $V^{T\times k\times 2}_i$ is the $i^{th}$ input sequence, $k$ is the number of keypoints, $L_{STG\mhyphen NF}$ is the trained STG-NF mapping network and $DL_i$ is the distance between each embedded pose sequence and $\mu_{Normal}$. 

Figure \ref{fig:pose_normal} shows the histogram of the distances of all embedded pose sequences from $\mu_{Normal}$. Figure \ref{fig:pose_normal_shang} demonstrates that the pose data is a discriminative feature in the ShanghaiTech. The distances of normal sequences in the validation and train sets are lower than those of anomalous data points in the validation set. This shows that the pose sequences are enriched enough to detect anomalies in the ShanghaiTech. However, this is different in the CHAD. Figure \ref{fig:pose_normal_chad} shows that using only pose data for CHAD is insufficient for detecting anomalies, as the embedded normal and anomalous pose sequences exhibit significant overlap. In other words, the trained model is not able to map the abnormal and normal pose sequences far enough in the latent space using pose data.

\section{Embedded Trajectory Distribution}
\label{sec:traj_normal}

\begin{table}[]
\centering
\caption{Results of STG-NF on ShanghaiTech \cite{lou2021ShanghaiTechAnomaly} and CHAD \cite{pazho2022chad}, trained using absolute and social trajectory formulation.}
\label{tab:stg_nf_traj_result}
\resizebox{\columnwidth}{!}{%
\begin{tabular}{c||cc||cc}
\rowcolor{Gray}
                       & \multicolumn{2}{c||}{ShanghaiTech \cite{lou2021ShanghaiTechAnomaly}}                                                                    & \multicolumn{2}{c}{CHAD \cite{pazho2022chad}}                                                                                          \\ \hline
                       \rowcolor{LightGray}
Formulation            & \multicolumn{1}{c|}{\begin{tabular}[c]{@{}c@{}}Absolute\\ Traj.\end{tabular}} & \begin{tabular}[c]{@{}c@{}}Social\\ Traj.\end{tabular} & \multicolumn{1}{c|}{\begin{tabular}[c]{@{}c@{}}Absolute\\ Traj.\end{tabular}} & \begin{tabular}[c]{@{}c@{}}Social\\ Traj.\end{tabular} \\ \hline \hline
AUC-ROC(\%) $\uparrow$ & \multicolumn{1}{c|}{64.5}                                                    & 64.3                                                 & \multicolumn{1}{c|}{61.2}                                                    & 60.1                                                  \\
AUC-PR(\%) $\uparrow$  & \multicolumn{1}{c|}{72.9}                                                    & 72.9                                                  & \multicolumn{1}{c|}{67.7}                                                    & 67.5                                                  \\
EER $\downarrow$       & \multicolumn{1}{c|}{0.38}                                                     & 0.40                                                   & \multicolumn{1}{c|}{0.42}                                                     & 0.43                                                  
\end{tabular}%
}
\end{table}
Pose data is an informative feature that can be used for anomaly detection. Still, other features, such as trajectory, may also provide valuable information for anomaly detection without the complexity of pose data.

\subsection{Absolute Trajectory}
Absolute trajectory refers to the movement of the center of a person in pixel space through $T$ frames. In this formulation, we gather trajectories of individuals available in the scene and feed this data to STG-NF \cite{stg_nf}. As it can be seen in Figure \ref{fig:formulation}, the center of a person is calculated by finding the middle point of the right hip and the left hip. We construct a spatiotemporal graph of the position of individuals and train STG-NF to find the function to map them to a latent space with a defined normal distribution. All the parameters are kept the same as the previous test in Section \ref{fig:pose_normal}. STG-NF was trained on both ShanghaiTech \cite{lou2021ShanghaiTechAnomaly}, and CHAD \cite{pazho2022chad} with the learning rate of $5e^{-4}$, the weight decay of $5e^{-5}$, and a batch size of 256 for eight epochs. Based on Table \ref{tab:stg_nf_traj_result}, in the Shanghaitech, pose information is a much more informative feature compared to trajectory. On CHAD, we do not observe significant improvement compared to the pose formulation. This shows detecting anomalous behavior on CHAD using trajectory is almost as difficult as using pose. For more insight, the distances between the embedded trajectories and the $\mu_{normal}$ are also calculated and depicted in Figure \ref{fig:traj_normal}. We use Equation \ref{eq:dl} for calculating distance. The only difference is that in this test $k$ is equal to one since we have one point per person. In the case of the ShanghaiTech, comparing Figure \ref{fig:pose_normal_shang} and \ref{fig:traj_normal_shang} reveals that losing pose data affects the quality of latent mapping since in Figure \ref{fig:traj_normal_shang}, the embedded train normal, validation normal, and validation anomalous data points are highly overlapping. Figure \ref{fig:traj_normal_chad} indicates that the model is confused and maps all the data points to the approximately same location in the latent space in CHAD. This shows that trajectory in isolation is not informative enough for anomaly detection.

\subsection{Social Trajectory}
\begin{figure*}[htbp]
\centering
\begin{subfigure}{0.48\textwidth}
  \centering
  \includegraphics[width=\linewidth]{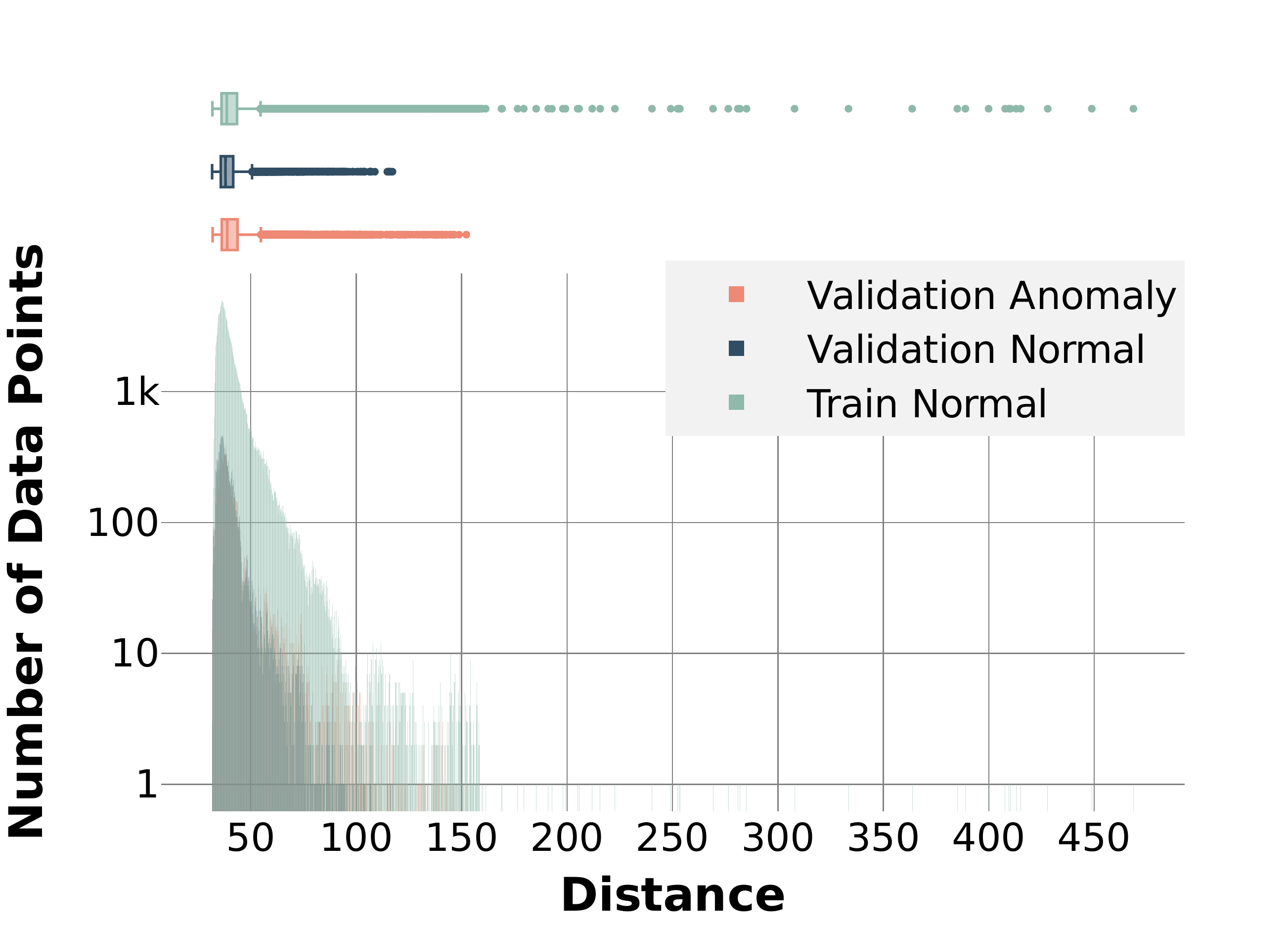}
  \caption{ShanghaiTech \cite{lou2021ShanghaiTechAnomaly}}
  \label{fig:traj_social_shang}
\end{subfigure}%
\begin{subfigure}{0.48\textwidth}
  \centering
  \includegraphics[width=\linewidth]{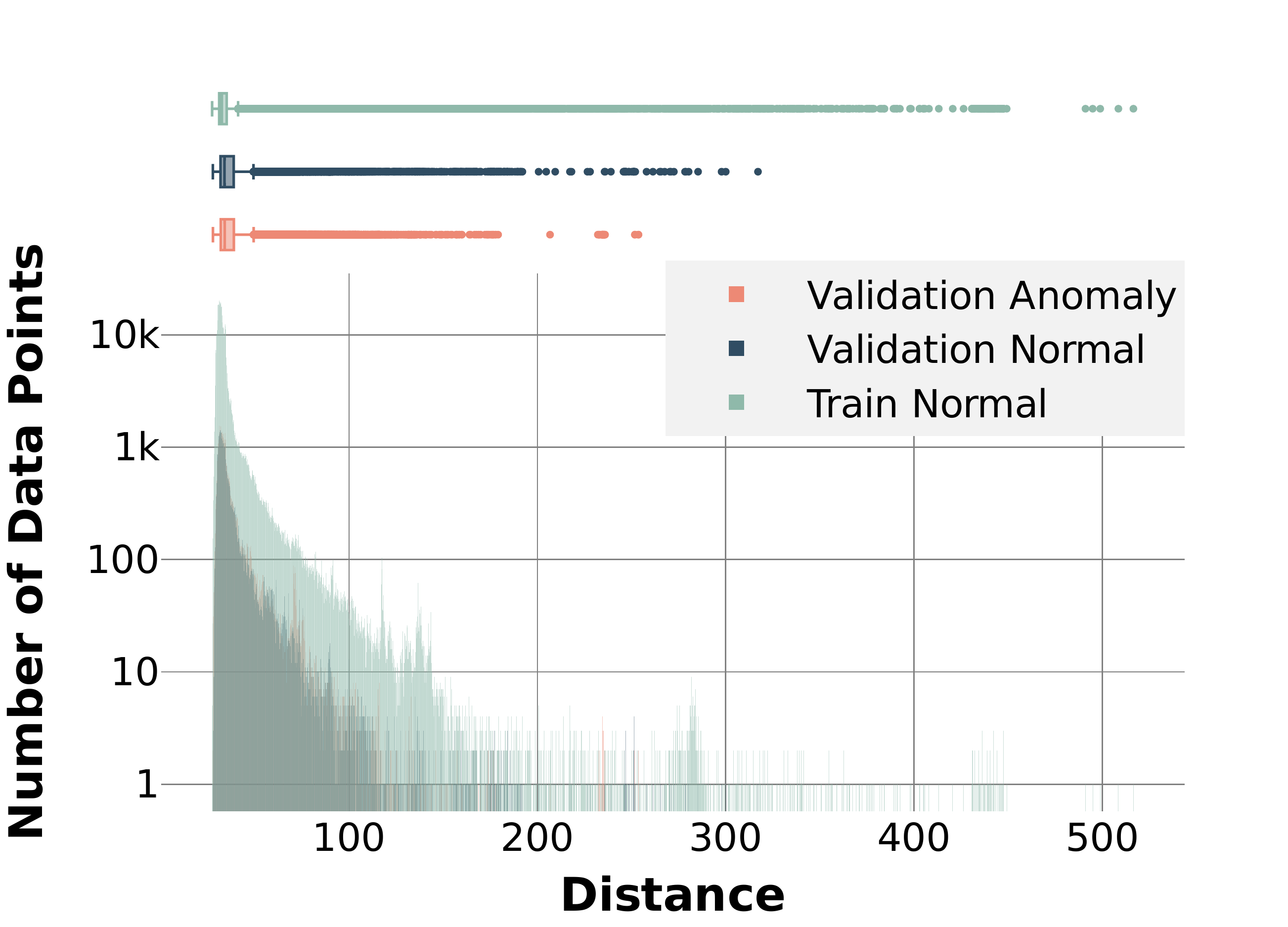}
  \caption{CHAD \cite{pazho2022chad}}
  \label{fig:traj_social_chad}
\end{subfigure}%
\vspace{-8pt}
 \caption{Histogram of the distances of all embedded social trajectories from the mean of the defined normal distribution ($\mu_{Normal}$) in the latent space. Detailed statistics can be found in Table \ref{tab:box}.}
\label{fig:traj_social}
\end{figure*}

Another way of integrating trajectory for anomaly detection is to consider the social interaction between people available in the scene and build the input spatiotemporal graph so that each node represents a person's location as depicted in Figure \ref{fig:formulation}. In this formulation, we chose to have a fully connected graph to remove predefined biases and let the model learn which social interactions between people in the scene are more valuable. The social interactions can give more insight into anomalies involving multiple people, such as people fighting with each other, etc. We zero-pad the input data to have a fixed number of nodes as the number of people in the scene varies. We chose 35 nodes since in both datasets the maximum number of people in a frame is less than 35. The input social trajectories and the latent space dimensions are subsequently changed to $24\times 35 \times 2$. For training, STG-NF \cite{stg_nf} on both ShanghaiTech \cite{lou2021ShanghaiTechAnomaly}, and CHAD \cite{pazho2022chad} datasets, we used the learning rate of $9e^{-4}$, the weight decay of $5e^{-5}$, and the batch size of 256 and trained the models for 12 epochs. The parameters of the latent normal distribution were kept the same. The results can be seen in Table \ref{tab:stg_nf_traj_result}. With this formulation of social trajectory, we almost get the same result compared to a single absolute trajectory test seen in Table \ref{tab:stg_nf_traj_result}. These findings imply that social trajectory is not helpful and in both ShanghaiTech and CHAD, the anomalous behavior examples involving multiple people are also discernible by looking at the individuals available in the scene. Figure \ref{fig:traj_social} depicts the distance between all data points and $\mu_{normal}$ on both datasets calculated by Equation \ref{eq:dl}. Based on Figure \ref{fig:traj_social_shang}, the mapping is not discriminative enough, and normal and anomalous data points are overlapping, similar to Figure \ref{fig:traj_normal_shang}. On the CHAD, Figure \ref{fig:traj_social_chad} reveals that the model is able to map the anomalous and normal data points to a different location at latent space, unlike Figure \ref{fig:traj_normal_chad}. This mapping is still not very discriminative, and normal and anomalous trajectories are heavily overlapping.

\section{Mean Distance Analysis}
\label{sec:mean_analysis}

\begin{figure*}[htbp]
\centering
\begin{subfigure}{0.48\textwidth}
  \centering
  \includegraphics[width=\linewidth]{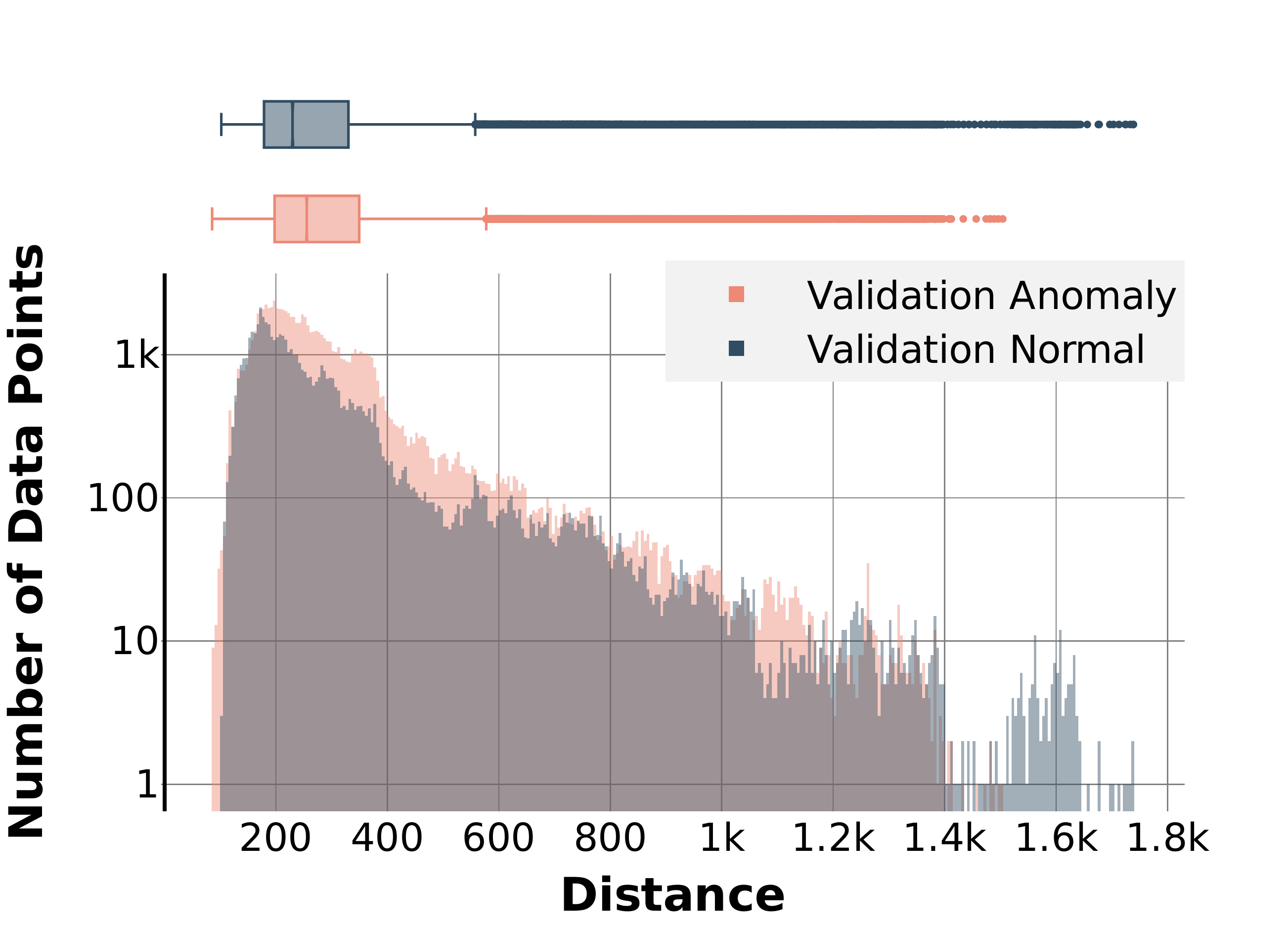}
  \caption{ShanghaiTech \cite{lou2021ShanghaiTechAnomaly}}
  \label{fig:pose_mean_shang}
\end{subfigure}%
\begin{subfigure}{0.48\textwidth}
  \centering
  \includegraphics[width=\linewidth]{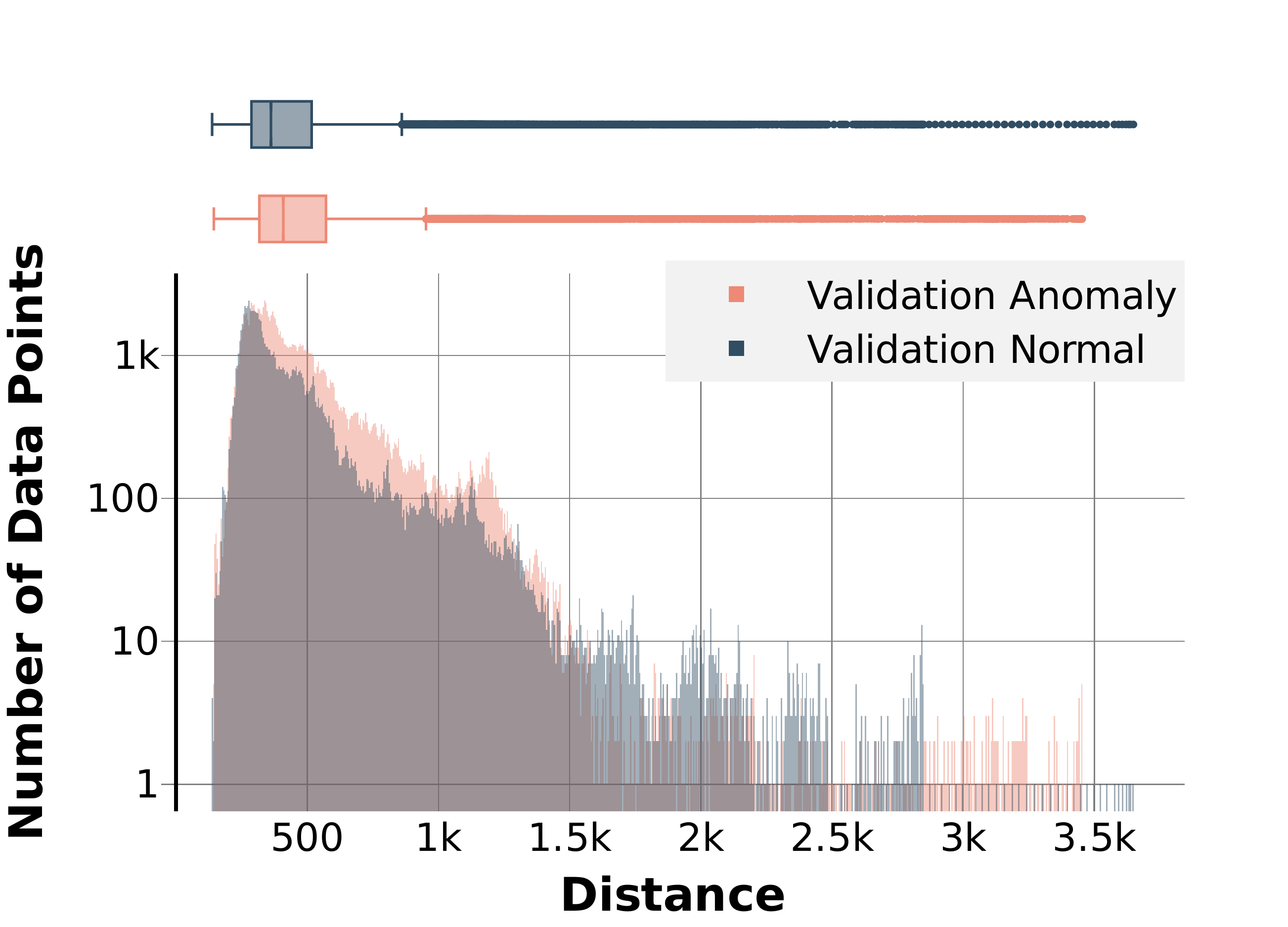}
  \caption{CHAD \cite{pazho2022chad}}
  \label{fig:pose_mean_chad}
\end{subfigure}%
\vspace{-8pt}
 \caption{Histogram of the distances of all pose sequences in the validation set from the mean of the normal pose sequences in the training set. Detailed statistics can be found in Table \ref{tab:box}.}
  \label{fig:pose_mean}
\end{figure*}

In this section, first, we introduce a metric called Signed Difference of Means ($S\mhyphen DoM$) to capture and quantify the inherent discriminative power of different input features (pose and trajectory). This data-driven methodology allows us to have an unbiased look at each dataset's characteristics without assuming any underlying distribution to limit our judgment. In the next step, we analyze ShanghaiTech and CHAD based on this new metric and discuss the characteristics of each dataset. 

\subsection{Signed Distance of Means}
\label{sec:sdom}

\begin{table}
\centering
\caption{This table summarizes $\Delta_n$ (distance between the mean of the training set and normal data points in the validation set), $\Delta_a$ (distance between the mean of the training set and anomalous data points in the validation set) and $S\mhyphen DoM$ on Shanghaitech \cite{lou2021ShanghaiTechAnomaly}, and CHAD \cite{pazho2022chad} datasets for pose and trajectory. }
\vspace{-10pt}
\label{tab:stat}
\begin{tabular}{c||cc||cc}
\rowcolor{Gray}
              & \multicolumn{2}{c||}{ShanghaiTech \cite{lou2021ShanghaiTechAnomaly}} & \multicolumn{2}{c}{CHAD \cite{pazho2022chad}}                   \\ \hline
\rowcolor{LightGray}
Type          & \multicolumn{1}{c|}{Pose}   & \multicolumn{1}{c||}{Traj.}  & \multicolumn{1}{c|}{Pose}  & \multicolumn{1}{c}{Traj.}  \\ \hline \hline
$\Delta_n$ $\downarrow$ & \multicolumn{1}{c|}{2.80}   & \multicolumn{1}{c||}{0.58}  & \multicolumn{1}{c|}{2.91}  & \multicolumn{1}{c}{0.30}  \\
$\Delta_a$ $\uparrow$ & \multicolumn{1}{c|}{3.75}   & \multicolumn{1}{c||}{0.76}  & \multicolumn{1}{c|}{2.76}  & \multicolumn{1}{c}{0.13}  \\ \hline
\rowcolor{LightGray}
$S\mhyphen DoM$ $\uparrow$ & \multicolumn{1}{c|}{0.95}   & \multicolumn{1}{c||}{0.18}    & \multicolumn{1}{c|}{-0.15} & \multicolumn{1}{c}{-0.17} 
\end{tabular}%
\end{table}

Our aim is to introduce a metric that can be used to quantify the discriminative power of different types of features in anomaly detection datasets. With this goal in mind, using Equation \ref{eq:mean}, we calculate $\mu_{tn}$, $\mu_{vn}$ and $\mu_{va}$ which represent the mean of normal sequences in the training set, the mean of normal sequences in the validation set and the mean of anomalous sequences in the validation set respectively.
\begin{equation}
\label{eq:mean}
    \mu^{T\times k \times 2} = \frac{1}{P}\sum_{i=0}^P V^{T\times k \times 2}_i
\end{equation}

In Equation \ref{eq:mean} $P$, $V$, $T$, and $k$ represent the number of data points, the sequence vector, the number of frames in the sequence, and the number of keypoints, respectively. The last dimension is two, which contains each keypoint's $(x ,y)$ position. Scaled Euclidean distances between means ($\Delta_n$ and $\Delta_a$) are defined as seen in Equation \ref{eq:dist_mean_n} and \ref{eq:dist_mean_a}. 
\begin{equation}
\label{eq:dist_mean_n}
    \Delta_n = \frac{1}{T} ||\mu_{tn}-\mu_{vn}||
\end{equation}
\begin{equation}
\label{eq:dist_mean_a}
    \Delta_a = \frac{1}{T} ||\mu_{tn}-\mu_{va}||
\end{equation}
Larger $\Delta_a$ shows the anomalous behavior in the validation set is drastically different from normal behavior in the training set. On the other hand, smaller $\Delta_n$ shows the normal behavior in training and validation sets are similar. Finally, for understanding how discriminative a feature is, we introduce a new metric called Signed Distance of Means ($S\mhyphen DoM$) described in Equation \ref{eq:sdom}. 

\begin{equation}
\label{eq:sdom}
  S\mhyphen DoM = \Delta_a - \Delta_n
\end{equation}

A larger $S\mhyphen DoM$ shows that anomalous datapoinst are more distinguishable, and the dataset is less challenging. 

\subsection{Pose Mean Distance}
\label{sec:pose_md}


\begin{table*}
\centering
\footnotesize
\caption{Box and whiskers plot details.}
\vspace{-9pt}
\label{tab:box}
\begin{tabular}{c||cccccc}
               & \multicolumn{3}{c||}{\cellcolor{DarkGray}\textbf{ShanghaiTech Campus Anomaly Dataset} \cite{lou2021ShanghaiTechAnomaly}}                                                                   & \multicolumn{3}{c}{\cellcolor{DarkGray}\textbf{CHAD Anomaly Dataset} \cite{pazho2022chad}}                                                              \\ \cline{2-7} 
               & \multicolumn{1}{c|}{\cellcolor{Gray}\textbf{Train Normal}} & \multicolumn{1}{c|}{\cellcolor{Gray}\textbf{Val. Normal}} & \multicolumn{1}{c||}{\cellcolor{Gray}\textbf{Val. Anomaly}} & \multicolumn{1}{c|}{\cellcolor{Gray}\textbf{Train Normal}} & \multicolumn{1}{c|}{\cellcolor{Gray}\textbf{Val. Normal}} & \cellcolor{Gray}\textbf{Val. Anomaly} \\ \cline{2-7} 
               & \multicolumn{6}{c}{\cellcolor{LightGray}\textbf{Embedded Pose Distribution }(Figure \ref{fig:pose_normal})}                                                                                                                                                                                                             \\ \hline
Lower Fence    & 23.13                                      & 23.44                                     & \multicolumn{1}{c||}{24.43}                 & 17.69                                      & 15.72                                     & 14.15                 \\
First Quartile & 28.32                                      & 28.18                                     & \multicolumn{1}{c||}{31.07}                 & 27.46                                      & 26.03                                     & 25.77                 \\
Median         & 29.94                                      & 29.72                                     & \multicolumn{1}{c||}{35.73}                 & 30.57                                      & 29.39                                     & 29.56                 \\
Third Quartile & 31.78                                      & 31.75                                     & \multicolumn{1}{c||}{47.52}                 & 33.97                                      & 32.91                                     & 33.51                 \\
Upper Fence    & 36.98                                      & 37.09                                     & \multicolumn{1}{c||}{72.1}                  & 43.74                                      & 43.22                                     & 45.12                 \\ \hline
        \cellcolor{LightGray}       & \multicolumn{6}{c}{\cellcolor{LightGray}\textbf{Embedde Absolute Trajectory }(Figure \ref{fig:traj_normal})}                                                                                                                                                                                              \\ \hline
Lower Fence    & 1.83                                       & 2.21                                      & \multicolumn{1}{c||}{2.39}                  & 6.52                                       & 6.3                                       & 6.31                  \\
First Quartile & 5.51                                       & 5.96                                      & \multicolumn{1}{c||}{6.43}                  & 6.97                                       & 7                                         & 7.01                  \\
Median         & 7.75                                       & 8.36                                      & \multicolumn{1}{c||}{9.15}                  & 7.1                                        & 7.18                                      & 7.2                   \\
Third Quartile & 10.27                                      & 11.19                                     & \multicolumn{1}{c||}{12.08}                 & 7.28                                       & 7.48                                      & 7.48                  \\
Upper Fence    & 17.42                                      & 19.03                                     & \multicolumn{1}{c||}{20.53}                 & 7.74                                       & 8.18                                      & 8.18                  \\ \hline
        \cellcolor{LightGray}       & \multicolumn{6}{c}{\cellcolor{LightGray}\textbf{Embedded Social Trajectory }(Figure \ref{fig:traj_social})}                                                                                                                                                                                                \\ \hline
Lower Fence    & 31.82                                      & 31.71                                     & \multicolumn{1}{c||}{31.96}                 & 27.25                                      & 27.63                                     & 27.7                  \\
First Quartile & 36.15                                      & 35.8                                      & \multicolumn{1}{c||}{36.28}                 & 30.96                                      & 31.68                                     & 31.81                 \\
Median         & 38.68                                      & 38.05                                     & \multicolumn{1}{c||}{38.84}                 & 32.42                                      & 33.82                                     & 33.92                 \\
Third Quartile & 43.55                                      & 41.75                                     & \multicolumn{1}{c||}{43.71}                 & 35                                         & 38.7                                      & 38.8                  \\
Upper Fence    & 54.66                                      & 50.67                                     & \multicolumn{1}{c||}{54.85}                 & 41.06                                      & 49.23                                     & 49.3                  \\ \hline
        \cellcolor{LightGray}       & \multicolumn{6}{c}{\cellcolor{LightGray}\textbf{Pose Mean Distance} (Figure \ref{fig:pose_mean})}                                                                                                                                                                                                               \\ \hline
Lower Fence    & -                                          & 102.03                                    & \multicolumn{1}{c||}{85.53}                 & -                                          & 137.18                                    & 143.88                \\
First Quartile & -                                          & 178.59                                    & \multicolumn{1}{c||}{197.56}                & -                                          & 287.27                                    & 317.05                \\
Median         & -                                          & 229.91                                    & \multicolumn{1}{c||}{255.34}                & -                                          & 361.51                                    & 408.39                \\
Third Quartile & -                                          & 330.22                                    & \multicolumn{1}{c||}{349.49}                & -                                          & 516.57                                    & 571.24                \\
Upper Fence    & -                                          & 557.63                                    & \multicolumn{1}{c||}{577.38}                & -                                          & 860.46                                    & 952.49                \\ \hline
        \cellcolor{LightGray}       & \multicolumn{6}{c}{\cellcolor{LightGray}\textbf{Absolute Trajectory Mean Distance} (Figure \ref{fig:traj_mean})}                                                                                                                                                                             \\ \hline
Lower Fence    & -                                          & 118.74                                    & \multicolumn{1}{c||}{106.95}                & -                                          & 310.12                                    & 274.29                \\
First Quartile & -                                          & 458.12                                    & \multicolumn{1}{c||}{399.15}                & -                                          & 751.53                                    & 788.47                \\
Median         & -                                          & 732.7                                     & \multicolumn{1}{c||}{698.66}                & -                                          & 1021.62                                   & 1137.94               \\
Third Quartile & -                                          & 1066.914                                  & \multicolumn{1}{c||}{1092.48}               & -                                          & 1752.02                                   & 1836.36               \\
Upper Fence    & -                                          & 1980.02                                   & \multicolumn{1}{c||}{2132.31}               & -                                          & 3252.65                                   & 3408.12               
\end{tabular}
\end{table*}


Replicating the experimental conditions of prior tests, we chose $T$ and stride equal to 24 and 6, respectively. After building the pose sequences, we perform a centralizing preprocessing step to eliminate the effect of the pose sequence's global position with respect to the boundaries of the frame. We take the first pose in the sequence and translate it to the center of the frame. However, for the following poses in the sequence, we use the same translation vector as the first pose to make sure the relative movements of a person through space are captured. Using metrics introduced in Section \ref{sec:sdom} we quantify the characteristics of the datasets as shown in Table \ref{tab:stat}. $S\mhyphen DoM$ on ShanghaiTech is equal to $0.95$ which shows pose data is a great discriminative feature. But on CHAD,  $S\mhyphen DoM$ of $-0.15$ indicates that pose data is not discriminative enough and CHAD is a much harder dataset compared to ShanghaiTech. 

\begin{figure*}[htbp]
\centering
\begin{subfigure}{0.48\textwidth}
  \centering
  \includegraphics[width=\linewidth]{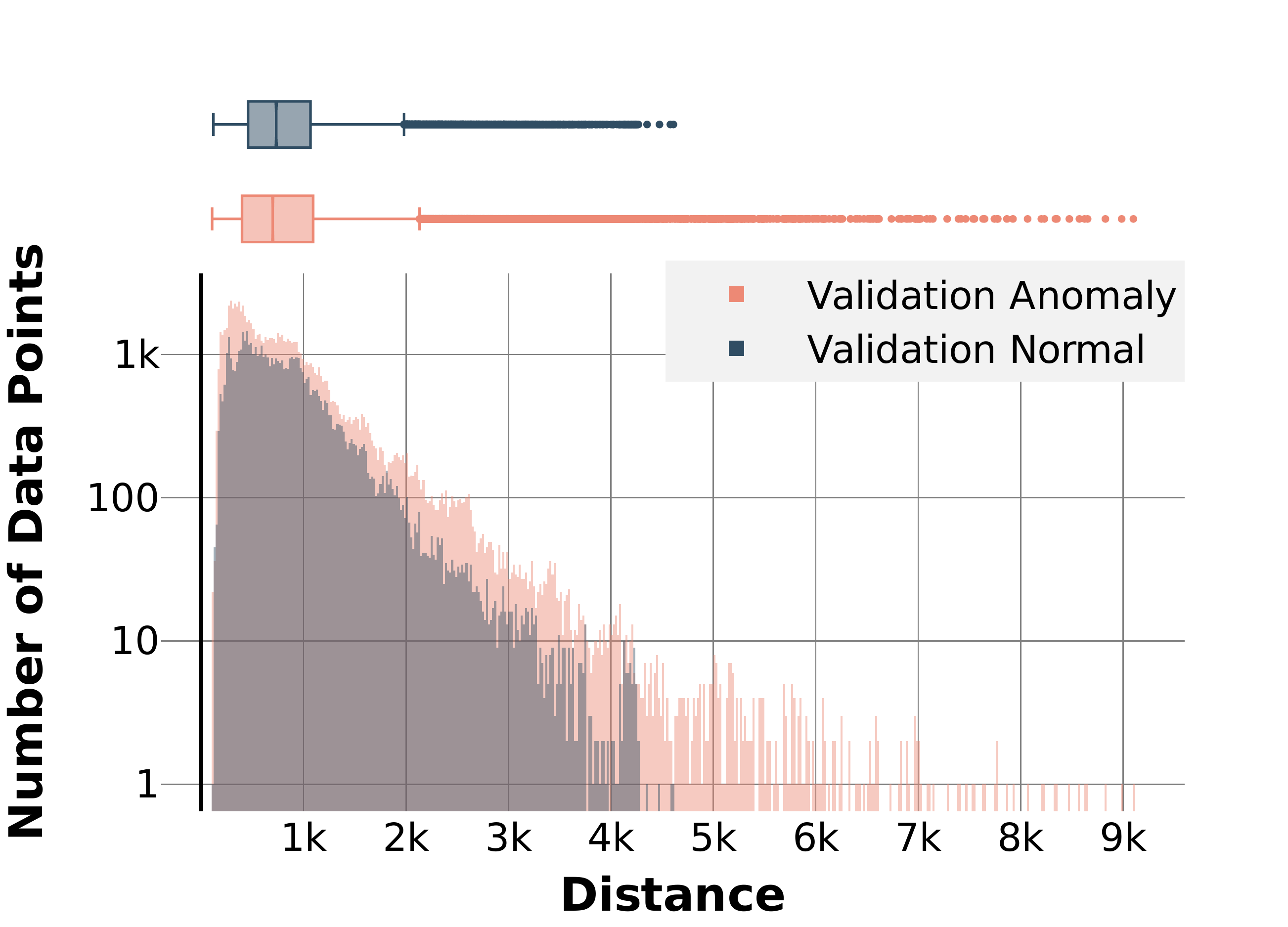}
  \caption{ShanghaiTech \cite{lou2021ShanghaiTechAnomaly}}
  \label{fig:traj_mean_shang}
\end{subfigure}%
\begin{subfigure}{0.48\textwidth}
  \centering
  \includegraphics[width=\linewidth]{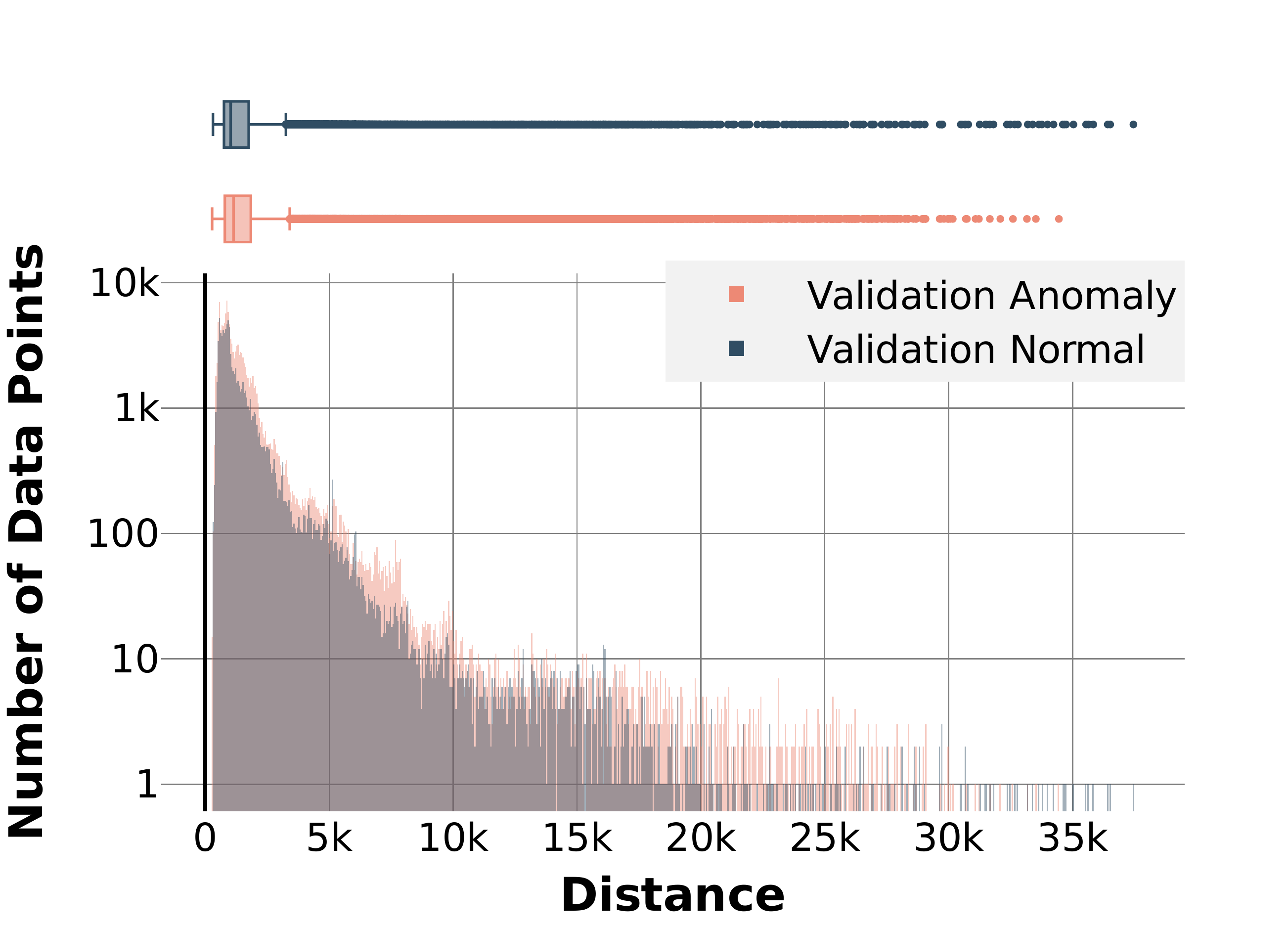}
  \caption{CHAD \cite{pazho2022chad}}
  \label{fig:traj_mean_chad}
\end{subfigure}%
\vspace{-8pt}
 \caption{Histogram of the distances of all trajectories in the validation set from the mean of the normal trajectories in the training set. Detailed statistics can be found in Table \ref{tab:box}.}
\label{fig:traj_mean}
\end{figure*}

For a more thorough examination, we also calculate the distance of each datapoint in the validation set from the $\mu_{tn}$ to visualize a histogram of distances and better understand the distribution of pose sequences using Equation \ref{eq:dis_stat}.

\begin{equation}
\label{eq:dis_stat}
    D_i = ||V^{T\times k \times 2}_i - \mu^{T\times k \times 2}_{tn}||
\end{equation}

Figure \ref{fig:pose_mean} shows that anomalous pose sequences in the validation set are relatively further from $\mu_{tn}$ than normal sequences in the validation set in both datasets and are somewhat distinctive.

\subsection{Absolute Trajectory Mean Distance}

We undertake the same experiment for analyzing the characteristics of absolute trajectory in ShanghaiTech \cite{lou2021ShanghaiTechAnomaly}, and CHAD \cite{pazho2022chad} datasets. We also centralize the trajectory with the same formulation used in Subsection \ref{sec:pose_md}. We report introduced metrics in Section \ref{sec:sdom} for trajectories in Table \ref{tab:stat}. In Shanghaitech, we can see that $S\mhyphen Dom$ for trajectory formulation is decreased by $0.77$ compared to the pose sequence formulation. This shows anomaly detection using only trajectory is much more challenging than using pose since the trajectory only has the information about how the center point of a person is moving. The same case happens in CHAD; $S\mhyphen Dom$ is diminished by $0.02$ compared to pose sequence formulation. Results also point out that CHAD seems to be a more challenging dataset for anomaly detection using both pose and trajectory.

For a more comprehensive analysis, we also plot the histogram of the distance between trajectories in the validation set and $\mu_{tn}$ using Equation \ref{eq:dis_stat} which can be seen in Figure \ref{fig:traj_mean}. As suggested by the low values of $S\mhyphen DoM$ seen in Table \ref{tab:stat}, we can see that normal and abnormal trajectories highly overlap in both datasets. This shows that the trajectory alone is not enriched enough to be used for anomaly detection in these datasets.

\section{Conclusion}
In this work, we analyze two anomaly detection datasets from multiple perspectives to better understand the challenges and obstacles of pose-based anomaly detection. We conduct extensive tests to evaluate the discriminative power of pose data. On top of that, we inspect trajectory with individual and social formulations to reveal whether it can be adopted for person anomaly detection or not. We introduce a new metric to further quantify the features of the dataset. Findings suggest in general pose has better discriminative power than trajectory. However, as per the results, both pose and trajectory might not be enough for challenging datasets. Taking advantage of the combination of the two, or an addition of other types of information can be beneficial for robustness and higher performances, especially in the case of real-world scenarios.

\section*{Acknowledgements}
This research is supported by the National Science Foundation (NSF) under Award No. 1831795.

{\small
\bibliographystyle{ieee_fullname}
\bibliography{egbib}
}

\end{document}